\documentclass[manuscript, screen]{acmart}
\AtBeginDocument{%
  }

\setcopyright{acmlicensed}
\copyrightyear{2018}
\acmYear{2018}
\acmDOI{XXXXXXX.XXXXXXX}
\acmConference[Conference acronym 'XX]{Make sure to enter the correct
  conference title from your rights confirmation email}{June 03--05,
  2018}{Woodstock, NY}
\acmISBN{978-1-4503-XXXX-X/2018/06}

\begin{document}

\title{Long-Tail Knowledge in Large Language Models: Taxonomy, Mechanisms, Interventions and Implications}

\author{Sanket Badhe}
\email{sanketbadhe@google.com}
\affiliation{%
  \institution{Google}
  \city{Mountain View}
  \state{California}
  \country{USA}
}

\author{Deep Shah}
\email{shahdeep@google.com}
\affiliation{%
  \institution{Google}
  \city{Mountain View}
  \state{California}
  \country{USA}
}

\author{Nehal Kathrotia}
\email{nehalk@google.com}
\affiliation{%
  \institution{Google}
  \city{Mountain View}
  \state{California}
  \country{USA}
}

\renewcommand{\shortauthors}{Badhe et al.}

\begin{abstract}

Large language models (LLMs) are trained on web-scale corpora that exhibit steep power-law distributions, in which the distribution of knowledge is highly long-tailed, with most appearing infrequently. While scaling has improved average-case performance, persistent failures on low-frequency, domain-specific, cultural, and temporal knowledge remain poorly characterized. This paper develops a structured taxonomy and analysis of long-Tail Knowledge in large language models, synthesizing prior work across technical and sociotechnical perspectives.

We introduce a structured analytical framework that synthesizes prior work across four complementary axes: how long-Tail Knowledge is defined, the mechanisms by which it is lost or distorted during training and inference, the technical interventions proposed to mitigate these failures, and the implications of these failures for fairness, accountability, transparency, and user trust. We further examine how existing evaluation practices obscure tail behavior and complicate accountability for rare but consequential failures. The paper concludes by identifying open challenges related to privacy, sustainability, and governance that constrain long-Tail Knowledge representation. Taken together, this paper provides a unifying conceptual framework for understanding how long-Tail Knowledge is defined, lost, evaluated, and manifested in deployed language model systems.

\end{abstract}

\ccsdesc[500]{Computing methodologies~Machine learning approaches}
\ccsdesc[500]{Computing methodologies}

\keywords{Long-Tail Knowledge, large language models, data sparsity, knowledge representation, evaluation practices, sociotechnical implications, Cultural concept popularity, Low-resource languages, Geopolitical bias, evaluation benchmarks}

\received{12 January 2026}
\received[revised]{12 March 2009}
\received[accepted]{5 June 2009}

\maketitle

\section{Introduction}

Large Language Models (LLMs) are increasingly used as interfaces for accessing and synthesizing information across a wide range of applications. Unlike traditional information retrieval systems that return documents based on explicit keyword matching, these models generate responses directly from internal parametric representations learned during training. This shift places greater importance on the statistical properties of the data used during pre-training. While scaling laws suggest that increasing model size and data volume yields predictable performance improvements \cite{kaplan2020scaling}, recent empirical work indicates that these gains are unevenly distributed across different types of knowledge. Improvements tend to concentrate on high-frequency facts and linguistic patterns, while knowledge that appears infrequently in training corpora exhibits weaker and less reliable retention \cite{kandpal2023struggle}.

The distribution of information in web-scale text corpora approximately follows a Zipfian or heavy-tailed power law, in which a small fraction of facts occur very frequently while the majority appear rarely \cite{chang2023speak}. We use the term Long-Tail Knowledge (LTK) to refer to this broad class of low-frequency information. This includes rare entities, niche domain expertise, low-resource languages, and historical or cultural facts with limited digital presence. There is no single agreed-upon definition of LTK in the literature. Prior work approaches this phenomenon from different perspectives, including frequency-based operationalizations in natural language processing and concerns about representational coverage and knowledge gaps in fairness and accountability research \cite{bender2021parrots, mallen2023trust}. This work adopts an inclusive scope in order to synthesize these perspectives into a unified analytical framework.

Across recent empirical studies, model performance is observed to degrade non-linearly as the frequency of the target knowledge decreases. Learning rare facts often requires disproportionately larger amounts of data compared to frequent ones \cite{kandpal2023struggle}. At the same time, standard training objectives prioritize minimizing loss on high-frequency tokens, which biases learning toward common patterns and reduces gradient signal for infrequent knowledge \cite{pezeshki2021gradient}. In practice, failures on long-tail queries often manifest as hallucinated responses. Rather than expressing uncertainty, models frequently generate fluent but factually incorrect statements when queried about rare or weakly represented topics \cite{azaria2023internal, ji2023hallucination}. These behaviors have been documented across domains, including settings involving languages and histories that are sparsely represented in training data \cite{joshi2025invisible}.

Research on LTK is distributed across multiple research communities. Analyses of memorization and training dynamics appear primarily in machine learning and systems venues \cite{tirumala2022memorization}. Work examining representational coverage and downstream impacts is more common in ethics and fairness research \cite{weidinger2021ethical}. Proposed mitigation strategies, such as retrieval-augmented generation and model editing, are typically introduced in natural language processing conferences \cite{lewis2020retrieval, meng2022rome}. Differences in terminology, benchmarks, and evaluation protocols across these communities make it difficult to compare findings or assess the scope of LTK limitations in a unified manner.

This paper addresses this fragmentation by introducing a structured conceptual framework that organizes existing work on LTK across definitions, mechanisms, evaluation practices, and sociotechnical implications.:

\begin{itemize}
    \item \textbf{RQ1 (Taxonomy):} What categories of LTK are defined or operationalized across different domains and tasks?
    \item \textbf{RQ2 (Mechanisms):} What training dynamics, architectural constraints, or inference-time processes are reported to contribute to the loss or fragile retention of rare knowledge?
    \item \textbf{RQ3 (Interventions):} What technical strategies have been proposed to improve LTK performance, and how are their tradeoffs characterized?
    \item \textbf{RQ4 (Implications):} What impacts of LTK gaps are reported in downstream applications, user interactions, or representational coverage?
\end{itemize}

Existing work on rare or underrepresented knowledge is dispersed across domains, tasks, and evaluation paradigms, often using incompatible definitions and measurement practices. This fragmentation makes it difficult to compare findings, identify recurring failure modes, or assess the real-world significance of reported results. As a consequence, LTK loss is frequently treated as an isolated technical limitation rather than a systemic property of contemporary model development and evaluation pipelines. We address a foundational gap in the literature by synthesizing fragmented empirical evidence on LTK in LLMs. This paper is the first to systematically synthesize these findings across definitions, mechanisms, mitigation strategies, and sociotechnical implicaxtions.

By organizing technical analyses of training dynamics alongside studies of downstream behavior and deployment contexts, this paper provides a structured account of how statistical sparsity in training data relates to observed limitations in model reliability, coverage, and consistency. This synthesis is necessary to support more rigorous evaluation practices, to clarify the limits of current mitigation strategies, and to inform responsible deployment decisions in settings where rare or specialized knowledge carries outsized social impact.

\begin{figure*}[t] 
    \centering
    \includegraphics[width=\textwidth]{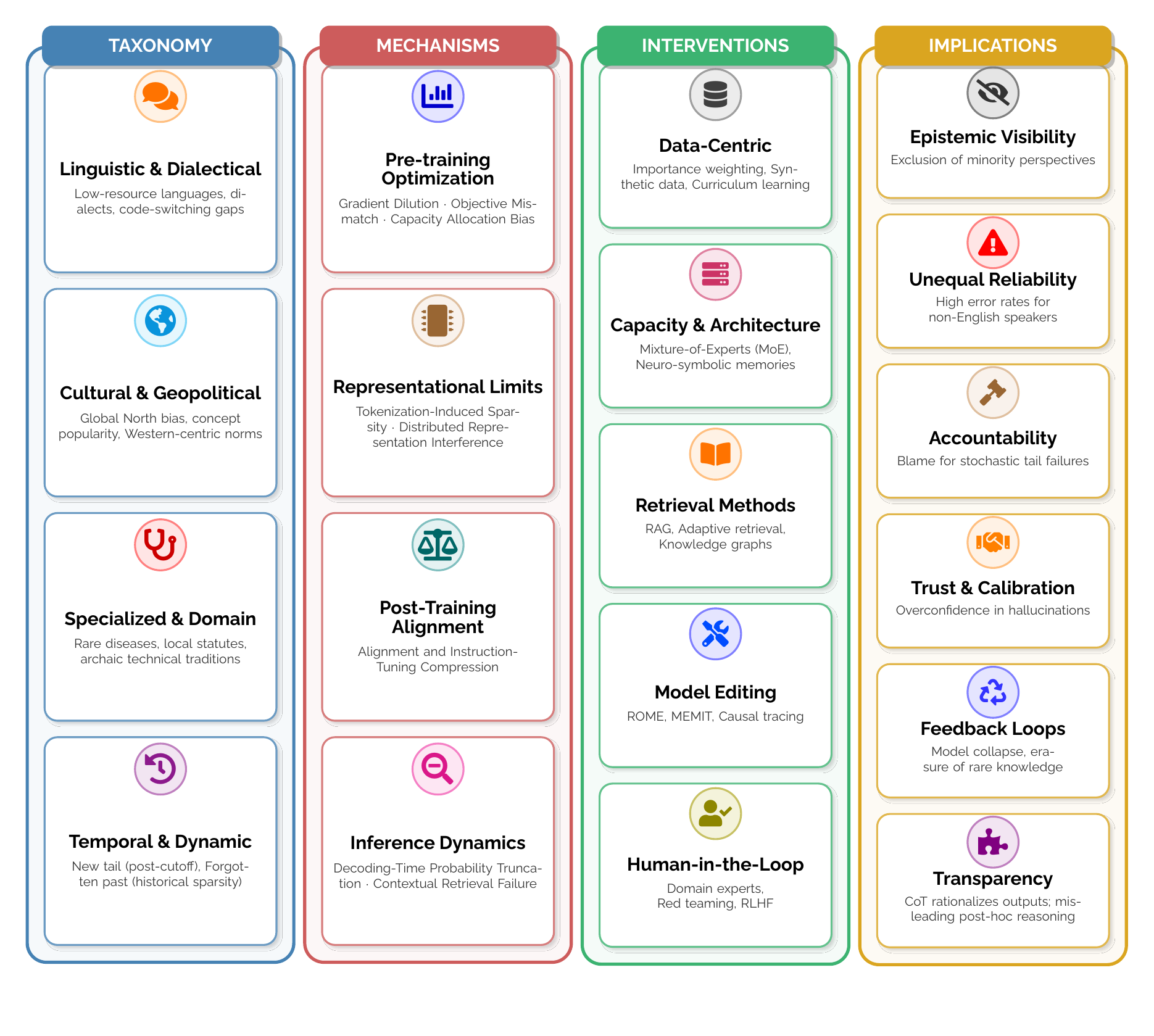}
    \caption{Overview of Long-Tail Knowledge in LLMs: Taxonomy, Mechanisms, Interventions, and Implications.}
    \label{fig:longtail_overview}
\end{figure*}

\section{The Taxonomy of the Long-Tail}

To address the LTK problem effectively, we must first categorize the types of information that systematically reside in regions of low empirical support within the training distribution. Across web-scale corpora, this sparsity is commonly measured through token, entity, or document frequency, and it follows a heavy-tailed distribution. Frequency-based rarity is therefore a necessary but insufficient condition for LTK. This section treats frequency as a shared underlying axis and introduces a taxonomy that categorizes LTK according to the properties that shape how sparsity manifests in practice.

Our analysis of the literature reveals four primary ontological dimensions of sparsity: linguistic, cultural, domain-specific, and temporal. These dimensions are not mutually exclusive and frequently interact. They collectively subsume the dominant ways in which LTK is operationalized and evaluated in empirical studies, including cases of compositional rarity and procedural or tacit knowledge that are weakly represented in declarative text corpora.

\subsection{Linguistic and Dialectical Sparsity}

The first dimension of the long tail is linguistic. While scaling laws hold for high-resource languages such as English, model performance degrades sharply for low-resource languages and non-standard dialects \cite{stanford2025mindgap}. This disparity is partly driven by tokenization artifacts. Tokenizers optimized for English frequently fragment words from low-resource languages into long sequences of subword units, diluting semantic coherence and weakening statistical learning signals \cite{petrov2023tokenization}. Research indicates that LLMs exhibit severe performance degradation on low-resource African languages, such as Amharic and Sepedi, where data scarcity is compounded by morphological complexity that standard tokenizers fail to capture \cite{adebara2025enhancing}.

Recent work highlights a distinct safety and reliability gap in this linguistic tail. Shen et al. \cite{wu2024lower} identify two failure modes for low-resource languages: a harmfulness curse, in which models are more likely to generate unsafe content, and a relevance curse, in which instruction-following and task adherence collapse. Linguistic sparsity also extends to dialectal variation within high-resource languages. The ReDial benchmark demonstrates that LLMs exhibit brittleness when processing African American Vernacular English and often fail on reasoning tasks that they solve reliably in standardized English \cite{hofmann2024dialects, groenwold2024one}. Code-switching presents a further challenge. Models frequently default to the dominant matrix language or fail to preserve syntactic consistency across language boundaries, reflecting compositional rarity rather than absence of component knowledge \cite{yong2023multilingual, hendy2023good}. Beyond dialects, The Script-Gap illustrates a critical safety blind spot in LLM-based systems: models exhibit consistent performance degradation when processing romanized messages compared to native scripts—a common practice in digital communication for Indian languages—even when the models appear to correctly infer the user's semantic intent. \cite{mani2025script}.  

\subsection{Cultural and Geopolitical Peripheries}

The second dimension concerns cultural grounding and geopolitical representation. LLMs exhibit a cultural concept popularity bias, in which entities and events salient to the Global North are recalled with high accuracy, while analogous concepts from underrepresented regions are omitted or hallucinated \cite{jiang2023cpopqa}. The CPopQA benchmark shows that models accurately rank holidays in the United States but perform substantially worse for culturally significant events in regions with lower digital visibility, including Sub-Saharan Africa and Southeast Asia \cite{jiang2023cpopqa, hershcovich2022challenges}.

This imbalance induces a Western-centric internal world model. Studies using benchmarks such as CANDLE and CAMeL demonstrate that LLMs struggle to reason about non-Western entities and frequently conflate distinct indigenous histories or project Western norms onto local contexts \cite{naous2024measuring, nguyen2023extracting}. When evaluating subjective global opinions, research indicates that LLMs—especially those fine-tuned with human feedback—systematically favor perspectives from Western, Educated, Industrialized, Rich, and Democratic (WEIRD) populations. Rather than being broadly representative, these models exhibit a substantial shift toward the views of liberal, educated, and wealthy demographics, while the opinions of many other indigenious groups remain significantly underrepresented. \cite{santurkar2023whose, durmus2023towards}. The challenge is not merely the absence of data but the Cultural Alignment problem, where the model's interpretative frame is incapable of processing the nuances of pluralistic value systems without explicit, resource-intensive adaptation \cite{li2025culturespa}.

\subsection{Specialized and Domain-Specific Tails}

The third dimension encompasses specialized and domain-specific knowledge that appears infrequently in general-purpose training corpora. This category includes both declarative facts and procedural or tacit knowledge, such as diagnostic workflows, legal procedures, and technical practices that are sparsely documented. In the medical domain, LLMs perform well on common conditions but exhibit severe deficits when reasoning about rare diseases \cite{chen2025mimic}. The MIMIC-RD study reports that open-source models fail to predict approximately 60\% of observed rare diseases even when relevant phenotypic information is explicitly provided in the prompt \cite{chen2025mimic}.

A parallel pattern is observed in the legal domain. While LLMs may demonstrate strong performance on standardized bar exam questions, they frequently hallucinate when queried about local statutes, procedural rules of lower courts, or infrequently cited case law \cite{dahl2024hallucinating, nay2024law}. This contrast highlights a persistent gap between transferable reasoning skills, which generalize from high-frequency patterns, and the retrieval or application of low-frequency, context-specific knowledge \cite{kandpal2023struggle}. Domain-specific tails also include archaic craftsmanship and specialized technical traditions, Scholars are increasingly applying LLMs to digitize hidden technical knowledge and archaic craftsmanship preserved in handwritten manuscripts. While recent work demonstrates that LLMs achieve near-human levels of accuracy when used for post-correction of transcriptions, evaluation remains fragile due to systematic temporal biases \cite{strickland2024unlocking, kanerva-etal-2025-ocr}. For instance, Levchenko identifies an over-historicization phenomenon where models insert archaic characters from incorrect historical periods rather than preserving original period-specific orthography \cite{zhang2025evaluating}.

\subsection{Temporal and Dynamic Knowledge}

The final dimension is temporal. LTK is not purely static but evolves with time and model training cutoffs. We distinguish between two temporal tails: the New Tail and the Forgotten Past. The New Tail comprises events and facts that emerge after a model’s training data cutoff. In these cases, models exhibit temporal misalignment, often hallucinating outdated information or failing to recognize recent changes \cite{luu2021time, lazaridou2021mind, kasai2022realtime}.

The Forgotten Past refers to historical knowledge with low contemporary digital presence. LLMs trained primarily on modern web text exhibit a recency bias, leading to degraded performance on historical facts that are infrequently discussed in current discourse \cite{dhingra2022time, jang2023temporal}. This degradation disproportionately affects regions and communities whose historical records were digitized late or unevenly, reinforcing existing gaps in representational coverage \cite{joshi2025invisible}.

\section{Mechanisms of Knowledge Loss}

The failure of LLMs to retain and retrieve LTK is not the result of a single technical bottleneck. It is a compound failure that cascades through every stage of the model pipeline. To structure our analysis of these failures, we categorize the mechanisms of knowledge loss into four distinct architectural levels: Pre-training Optimization, Representational Constraints, Post-Training Alignment, and Inference Dynamics.

\textbf{Summary of Architectural Failure Modes:}
\begin{itemize}
    \item \textbf{Level 1: Pre-training Optimization.} The statistical nature of the loss function prioritizes high-frequency patterns. This leads to gradient dilution for rare facts and a capacity allocation bias that favors generic heuristics over specific details.
    \item \textbf{Level 2: Representational Constraints.} The discrete nature of tokenization and the phenomenon of superposition create structural barriers. These barriers prevent the coherent encoding of rare entities and cause distributed interference in the parameter space.
    \item \textbf{Level 3: Post-Training Alignment.} Reinforcement learning interventions introduce an alignment tax. This encourages models to hedge or abstain from answering tail queries to minimize the risk of hallucination or unsafety.
    \item \textbf{Level 4: Inference Dynamics.} Standard decoding strategies truncate the low-probability tail of the distribution. This systematically filters out correct but rare tokens during generation.
\end{itemize}

\subsection{Level 1: Pre-training Optimization and Objectives}

The primary driver of tail loss is the optimization landscape of the pre-training phase. We identify three specific mechanisms that degrade the retention of low-frequency data.

\subsubsection{Gradient Dilution}
In standard autoregressive training, the model minimizes the negative log-likelihood of the next token. This objective function is inherently frequency-dependent. For a tail fact $f$ with probability $p(f)$ in the training corpus, the expected gradient update $\mathbb{E}[\nabla \mathcal{L}]$ is proportional to $p(f)$. Consequently, high-frequency concepts generate consistent and high-magnitude gradient signals that stabilize the weights of the model. In contrast, long-tail facts produce sparse and high-variance updates. These updates are frequently overwritten by the noise of subsequent updates from dominant data \cite{pezeshki2021gradient}. This phenomenon is known as Gradient Starvation. It results in a model that minimizes loss on the head of the distribution while leaving tail features under-learned \cite{chen2022gradtail, kandpal2023struggle}.

\subsubsection{Objective Mismatch and the Hallucination Nexus}
A fundamental misalignment exists between the Maximum Likelihood Estimation objective and the goal of factual retention. The objective incentivizes the model to minimize perplexity rather than maximize factual accuracy. For a rare fact, the model can often achieve a lower loss by predicting a high-probability generic continuation rather than the low-probability true entity \cite{mallen2023trust}. This creates a pressure to learn heuristic associations. For example, the model may predict that a person born in 18th century France is a peasant rather than a specific individual. This statistical correlation drives the Hallucination Nexus. There is a strong inverse correlation between the frequency of a fact in the training data and the hallucination rate of the model. The low signal-to-noise ratio in tail activations triggers the generation of high-frequency priors that are statistically likely but factually incorrect \cite{zhang2023siren, ji2023survey, azaria2023internal}.

\subsubsection{Capacity Allocation Bias}
Transformers exhibit a Simplicity Bias where the model's inductive bias and optimization objectives preferentially favor low-sensitivity functions or simple rules over the memorization of high-complexity sparse data. Bhattamishra et al. \cite{bhattamishra2023simplicity} demonstrate that this bias leads Transformers to converge to simple hypotheses, which aids generalization but can impair the learning of complex tail information. In the context of LLMs, Barron \& White \cite{nezhurina2025too} show that limited capacity forces models to prioritize rule extraction over factual memorization, whereas larger models utilize their parameter budget to store factual mappings at the expense of rule-based extrapolation. Tirumala et al. \cite{tirumala2022memorization} further note that while models memorize the head (nouns and numbers) first, larger scales actually mitigate the loss of LTK rather than compressing it into noisy representations.

\subsection{Level 2: Representational Constraints}

Even when gradients are sufficient, structural constraints in the architecture of the Transformer can prevent the accurate storage of tail knowledge.

\subsubsection{Tokenization-Induced Sparsity}
The discrete input layer imposes a structural bottleneck known as Tokenization-Induced Sparsity. Standard Byte-Pair Encoding or Unigram tokenizers are optimized to minimize the sequence length of the training corpus. This effectively compresses common English words into single tokens. Conversely, rare entities and low-resource language terms are fragmented into long sequences of sub-word units \cite{petrov2023tokenization, sennrich2015neural}. Wang et al. \cite{ali2024tokenization} demonstrate that this fragmentation often leads to incorrect tokenization that is misaligned with human comprehension, hindering the model's ability to understand input precisely and causing nonsensical responses. While the attention mechanism can theoretically learn the compositionality of these fragments, Kudo \cite{kudo2018subword} identifies that treating these varied sequences as completely different inputs creates a spurious ambiguity that can degrade the model's robustness and increase error rates.

\subsubsection{Distributed Representation Interference}
LLMs operate in a regime of superposition where the model represents more features than it has dimensions by storing them non-orthogonally. While this allows for high capacity, it introduces Distributed Representation Interference \cite{liu2025superposition}. Frequent patterns occupy the principal directions of the representation space. This forces rare facts to be encoded in the noisy residuals \cite{elhage2022toy}. Recent theoretical work by Liu et al. \cite{henighan2025superposition} demonstrates that in this strong superposition regime,  the interference penalty is disproportionately borne by low-frequency features, which are squeezed into representations with high overlaps. These features are easily overwritten by the crosstalk from high-frequency features that share the same polysemantic neurons \cite{ bills2023language}.

\subsection{Level 3: Post-Training Alignment}

The fine-tuning phase introduces distinct failure modes that are not present in the base model.

\subsubsection{Alignment and Instruction-Tuning Compression}
Post-training interventions such as Reinforcement Learning from Human Feedback introduce an Alignment Tax on tail knowledge. RLHF typically optimizes for helpfulness and safety using a reward model trained on a small and high-quality dataset that is heavily biased towards the head of the distribution \cite{casper2023open, xiao2025algorithmicbiasaligninglarge, ouyang2022training}. This process induces a form of mode collapse. The model learns to hedge or abstain from answering obscure questions to avoid the penalty of being incorrect. Alternatively, it converges to a generic response style that suppresses specific but rare details \cite{bai2022training, kirk2024understanding}. Empirical studies show that heavy instruction tuning can degrade the calibration of the model on long-tail facts. The model over-generalizes safety refusals to obscure but harmless queries \cite{lin2024mitigating, wei2023instruction}.

\subsection{Level 4: Inference Dynamics}

Finally, even if knowledge is retained in the weights, it may be inaccessible during the generation phase.

\subsubsection{Decoding-Time Probability Truncation}
Standard decoding strategies often prune tail knowledge during inference. Nucleus sampling (top-p) and top-k sampling explicitly truncate the tail of the probability distribution to ensure coherence and diversity \cite{holtzman2019curious}. However, the correct token for a long-tail fact often resides in this truncated region of low probability. By strictly filtering for high-probability tokens, these decoding algorithms systematically silence the correct but rare answers \cite{wang-etal-2024-factuality}. This forces the model to select a more common but incorrect alternative from the head of the distribution \cite{meister2020beam}.

\subsubsection{Contextual Retrieval Failure}
Rare facts often have weak attention keys that are easily overshadowed by stronger context signals. This leads to Contextual Retrieval Failure. Research indicates that activating these fragile memories requires prompts of high specificity that mimic the exact context seen during training \cite{mallen2023trust}. When the prompt deviates even slightly, the internal retrieval mechanism fails to attend to the relevant parameter subspace. This leads to a silently known fact that is inaccessible during generation \cite{lewis2020retrieval, joshi2024retrieval}.

\section{Survey of Mitigation Strategies}

We now examine the strategies proposed to address LTK in LLMs. This section categorizes interventions by their point of operation within the model lifecycle: data curation, architectural design, inference-time retrieval, parameter editing, and human oversight. We document the operational mechanisms of each approach alongside their empirically observed trade-offs and limitations.

\subsection{Data-Centric Interventions}
The most direct approach involves modifying the training data distribution to amplify the signal of rare concepts. One prominent strategy is importance weighting. This method adjusts the loss function to penalize errors on tail examples more heavily than on head examples. Techniques such as GradTail dynamically increase gradient weights for rare tokens which forces the model to allocate more capacity to learning them \cite{chen2022gradtail, pezeshki2021gradient}. While effective in controlled settings, this approach risks catastrophic forgetting of the head as the optimization landscape becomes distorted.

Another data-centric method is the generation of synthetic data to populate the tail. Frameworks such as LLM-AutoDA leverage stronger models to generate diverse and high-quality training examples for under-represented classes. This effectively flattens the Zipfian distribution artificially \cite{wang2024llmautoda}. Similarly, curriculum learning strategies demonstrate that training on high-quality synthetic textbooks can improve coverage of niche domains compared to raw web scrapes \cite{gunasekar2023textbooks}. However, reliance on synthetic data introduces the risk of model collapse. Recursive training on generated data leads to a loss of variance and the erasure of the true tail over time \cite{shumailov2024curse}.

\subsection{Capacity and Architectural Allocation}
Architectural interventions seek to scale the memory capacity of the model without incurring prohibitive computational costs. The Mixture-of-Experts paradigm replaces dense layers with a set of specialized expert networks that are activated sparsely via a routing mechanism. Models like Mixtral 8x7B demonstrate that this approach allows for a massive increase in total parameters—providing 47B total parameters while only activating 13B per token—thereby creating vast potential memory slots for long-tail facts while maintaining efficient inference \cite{jiang2024mixtral, fedus2022switch}. However, recent analysis reveals that routing mechanisms often collapse to a few super experts that handle the majority of tokens. This leaves the tail-specific experts under-utilized and undertrained \cite{dai2024unveiling, li2024meid}.

Other architectural proposals include the integration of explicit memory layers or k-nearest neighbor components directly into the transformer blocks. These neuro-symbolic hybrids attempt to decouple factual storage from reasoning which allows the model to query an internal key-value store for rare facts \cite{khandelwal2020generalization}. While promising, these methods often suffer from high latency and the challenge of keeping the memory store synchronized with the evolving weights of the main network \cite{mallen2023trust}.

\subsection{Retrieval and Externalization-Based Methods}
Retrieval-Augmented Generation externalizes LTK by moving the burden of memorization from the weights of the model to a searchable index. By retrieving relevant documents at inference time, RAG allows the model to access a vast and updateable corpus of tail facts without retraining \cite{lewis2020rag, guu2020retrieval}. This approach is particularly effective for new tail knowledge that emerges after the training cutoff \cite{vu2023freshllms}.

However, RAG has significant limitations. Its effectiveness depends entirely on the quality of the retrieval system which itself may suffer from bias towards head documents \cite{zhang2024role}. Furthermore, standard RAG pipelines apply retrieval indiscriminately. This often hurts performance on queries where the internal parametric memory of the model is sufficient. To address this, adaptive retrieval methods use metrics such as Generative Expected Calibration Error to detect when a query falls into the long tail and trigger retrieval only when necessary \cite{zhang2024role, mallen2023trust}. Despite these refinements, RAG remains limited by the context window bottleneck and the ability of the model to reason over conflicting retrieved information \cite{weller2025theoreticallimitationsembeddingbasedretrieval, shao2023enhancing, gao2024survey}.

\subsection{Model Editing and Localized Updates}
For specific errors in the tail, model editing techniques offer a way to surgically modify parameters. Methods such as ROME and MEMIT use causal tracing to identify the specific neurons encoding a fact and update them via a closed-form solution \cite{meng2022rome, meng2023memit}. This allows for the correction of outdated or incorrect tail facts without the cost of full fine-tuning.

While powerful for single-point edits, these methods face significant scalability challenges. Research indicates that sequential edits can accumulate ripple effects where updates to one fact inadvertently damage the representations of related entities or degrade the general reasoning capabilities of the model \cite{cohen2023evaluating, gu2024model}. Furthermore, lifelong editing eventually leads to a degradation of the internal coherence of the model which renders it unusable after a certain number of updates \cite{yao2023editing, hu2024wilke}.

\subsection{Human-in-the-Loop and Institutional Interventions}
Sociotechnical interventions focus on integrating human oversight to verify and correct tail outputs. Human-in-the-Loop frameworks deploy domain experts to audit model outputs in high-stakes fields such as law and medicine. This process reduce hallucinations and increase safety before they reach the end user \cite{AuditLLM}. Reinforcement Learning from Human Feedback (RLHF) attempts to align the model with human preferences, suffer from algorithmic bias due to optimization regularization — a formal mechanism by which minority preferences are effectively disregarded as preference collapse while dominant human preferences get amplified during alignment \cite{xiao2025algorithmicbiasaligninglarge, casper2023open}. Barman et al. \cite{barman2025pluralistic} argue that this reliance on potentially homogeneous evaluator groups can lead to sycophancy, where the model mirrors the one-sided opinions or ideologies of its tutors, often at the expense of scientific adequacy or nuanced tail knowledge.

Institutional interventions also include the development of red teaming protocols specifically designed to probe for long-tail failures. However, these human-centric approaches are inherently unscalable. They rely on the availability of experts who possess the rare knowledge in question. This creates a bottleneck that prevents broad coverage of the long tail \cite{ganguli2022red}. To mitigate these constraints, some suggest shifting toward pluralistic feedback panels that represent a wider spectrum of epistemic standpoints, aiming to achieve interactive objectivity rather than simple preference averaging \cite{barman2025pluralistic}.

\section{Sociotechnical Implications}

The technical issues discussed in the previous section lead to specific problems when LLMs are used in real-world systems. This section looks at the evidence to show how failures in LTK create systemic risks. We organize this analysis into seven parts, ranging from whose knowledge is included to how we audit these systems.

\subsection{Epistemic Visibility and Knowledge Inclusion}
Studies show that LLMs filter information in a way that favors dominant ideas. Prior work on computational mediation argues that large-scale automated systems do not merely retrieve facts but actively shape epistemic authority by privileging dominant representations \cite{gillespie2018custodians}. Recent analyses of LLMs show that this effect persists in generative systems, where scale and fluency create an appearance of objectivity while systematically omitting minority and low-frequency perspectives \cite{bender2021parrots, plum2025identityawarelargelanguagemodels}. This exclusion can be measured. Kandpal et al. \cite{kandpal2023struggle} show that the ability of a model to answer a question depends directly on how many times that fact appears in the training data. This makes information with low digital presence almost invisible to the model. This creates a cycle where the model boosts already popular ideas while pushing down local knowledge that does not have enough data to be memorized.

\subsection{Unequal Reliability Across User Populations}
The reliability of these models depends heavily on who is using them. Research shows a digital divide where speakers of languages with less data face more errors. A study by the Stanford HAI policy group found that LLMs trained mostly on English data have higher error rates and are less safe when used in languages like Swahili or Burmese \cite{stanford2025mindgap}. This gap also extends to cultural and regional knowledge. Myung et al. \cite{lee2024blend} utilized the BLEnD benchmark to evaluate everyday knowledge across diverse cultures and found a massive disparity: while models achieved high accuracy on questions about United States culture, performance dropped for Ethiopian culture questions prompted in Amharic. Similarly, users in the Global South seeking information about local entities often receive answers grounded in Western perspectives or pure hallucination, as demonstrated by benchmarks like CAMeL which reveal significant alignment failures for Arab cultural contexts compared to Western ones \cite{alhafni2024investigating, jiang2023cpopqa}.

\subsection{Accountability Ambiguity in AI-Assisted Decision-Making}
Long-tail failures complicate the attribution of responsibility in AI-assisted workflows. When a traditional system fails due to a clear logic error, the fault is often assignable. However, failures rooted in missing knowledge occupy a gray zone. Dahl et al. \cite{dahl2024hallucinating} document how legal professionals have faced sanctions for submitting hallucinated citations generated by LLMs. These failures often arise not because the model breaks in a traditional sense but because it confidently fills knowledge gaps with statistically plausible but non-existent cases \cite{ho2025free}. This creates an accountability vacuum where developers may claim the model is working as intended by minimizing perplexity while users are unable to verify the obscure information that led to the error \cite{raji2022fallacy, widder2023limits}. The result is a system where responsibility for tail failures is difficult to locate as the error stems from the probabilistic nature of the model itself rather than a specific bug \cite{ganguli2022red}.

\subsection{Trust Calibration and Overconfidence Effects}
A critical sociotechnical risk is the misalignment between model confidence and factual accuracy in the tail. LLMs do not exhibit epistemic humility and often present hallucinations with the same rhetorical fluency as verified facts. Zhang et al. \cite{zhang2025law} propose the Log-Linear Law of Hallucination which predicts that error rates increase linearly as knowledge popularity decreases yet model confidence does not degrade proportionally. This calibration failure leads to user over-reliance. Studies on faithfulness show that users frequently accept coherent but incorrect answers in specialized domains because the tone of the model mimics authoritative expert discourse \cite{lin2022truthfulqa, turpin2023unfaithful}. This is particularly dangerous in long-tail scenarios where users turn to the model precisely because they lack the knowledge to verify the answer themselves \cite{mallen2023trust, ji2023survey}.

\subsection{Feedback Loops and Knowledge Marginalization Over Time}
The integration of LLM outputs into the web creates recursive feedback loops that can degrade the quality of public knowledge. Shumailov et al. \cite{shumailov2024aimc} define model collapse as a degenerative process where models trained on synthetic data progressively lose variance and converge on the mode of the distribution. This process disproportionately erases the tails of the distribution. As synthetic content floods the web, rare facts are replaced by generic approximations. This makes them even less likely to be learned by future iterations of the model. Empirical work suggests that even with verification, the accumulation of synthetic data can lead to the permanent loss of low-frequency concepts from the digital record and reinforces the hegemony of the head over time \cite{alemohammad2024self}.

\subsection{Limits of Transparency and Explainability for Rare Knowledge}
Explainability mechanisms often fail to faithfully reflect internal model states. Turpin et al. \cite{turpin2023unfaithful} demonstrate that Chain-of-Thought (CoT) reasoning can be systematically unfaithful: models may generate plausible reasoning chains that rationalize an output rather than reveal the causal basis of the decision. While this limitation applies broadly, it is particularly consequential for long-tail knowledge. In head regimes, incorrect reasoning can often be detected through redundancy, consensus, or external verification. In contrast, long-tail failures lack such safeguards, making post-hoc explanations especially misleading when the model has no robust internal representation of the queried knowledge. Prior work shows that explanation faithfulness and uncertainty signaling are limited, providing little assurance that a correct-looking explanation corresponds to genuine knowledge rather than fragile inference \cite{lanham2023measuring, kadavath2022language}.

\section{Critical Analysis of Evaluation and Accountability}

The structural invisibility of LTK is reinforced by the current paradigm of model evaluation. While benchmarks serve as the primary mechanism for establishing trust and verifying capabilities, our analysis reveals that they often systematically exclude the long tail, thereby producing an incomplete picture of model reliability. This section examines how these measurement gaps distort accountability claims and complicate the safe deployment of LLMs in high-stakes domains.

\subsection{Scope Limitations of Benchmark-Based Evaluation}
Standard benchmarks such as MMLU (Massive Multitask Language Understanding) and BIG-bench have become the de facto standards for assessing model quality. However, empirical analysis suggests that these suites are heavily biased towards head knowledge. The construction of these benchmarks often relies on scraping questions from standardized human exams or Wikipedia-derived trivia, sources that naturally reflect the dominant information distribution of the web \cite{hendrycks2021mmlu, srivastava2022bigbench}. Consequently, facts that reside in the statistical tail such as local histories, indigenous botanical knowledge, or niche technical specifications are functionally absent from the evaluation set \cite{dodge-etal-2021-documenting, liang2023holistic}.

\subsection{Metric Aggregation and the Visibility of Tail Failures}
The prevailing practice of reporting model performance via single aggregate metrics actively conceals long-tail failures. Statistical aggregation operates as a smoothing function that drowns out the signal of rare errors. Oakden-Rayner et al. \cite{oakden2020hidden} formally define this phenomenon as hidden stratification, demonstrating that a model can achieve state-of-the-art aggregate performance while consistently failing on clinically meaningful minority subsets. For instance, in medical imaging, a high overall accuracy can mask the model's consistent inability to detect a rare but aggressive cancer subtype.

To counteract this, researchers have proposed slice-based evaluation, which partitions test sets by entity frequency, dialect, or demographic attribute. Studies utilizing these granular metrics reveal that worst-group accuracy often diverges sharply from average performance \cite{buolamwini2018gender, choudhury2024designdisaggregated}. However, standard industry leaderboards rarely report these disaggregated statistics, privileging a monolithic view of performance that is calibrated to the statistical majority \cite{blodgett2020language}.

\subsection{Accountability Claims Under Partial Measurement}
The gap between what is measured and what is claimed creates a crisis of accountability. When developers claim a model is safe or reliable based on head-heavy benchmarks, they are making a universal claim supported only by partial evidence. This disconnect complicates the assignment of responsibility when failures occur in the tail. Researcher characterize this as the Fallacy of AI Functionality, where internal validity such as benchmark performance is mistakenly equated with external validity i.e. real-world reliability \cite{raji2022fallacy}.

Empirical audits show that the opacity of closed model evaluations exacerbates this issue. Without access to the specific prompts and test sets used by developers, independent auditors cannot verify whether a claimed capability extends to the long tail or is strictly limited to the tested head examples \cite{widder2023limits}. This lack of auditability allows developers to disclaim liability for unforeseen errors, even when those errors are predictable consequences of frequency-based learning dynamics \cite{bommasani2021opportunities, AIaccountability}.

\subsection{Evaluation Practices in Regulated and High-Stakes Domains}
In domains such as law and medicine, the stakes of long-tail failure are non-negotiable. However, current evaluation practices often fail to align with professional standards of care. In the legal field, models are often evaluated on their ability to pass the Uniform Bar Exam (UBE). While GPT-4 has demonstrated a 90th percentile score on the UBE, this metric assesses general legal reasoning, not the retrieval of specific, low-frequency case law \cite{katz2024gpt4}. Deployment audits reveal that these same models frequently hallucinate citations when dealing with obscure or local jurisdictions, a failure mode not captured by the UBE but critical for malpractice liability \cite{nay2024law}.

Similarly, in healthcare, the focus on USMLE (United States Medical Licensing Examination) performance prioritizes textbook knowledge over clinical judgment in rare scenarios. Systematic reviews of medical LLMs find that while models excel at answering multiple-choice questions, they struggle with assessments that simulate the ambiguity and sparsity of real-world rare disease diagnosis \cite{nori2023capabilities, singhal2023expert, RareBench}. This mismatch indicates that regulatory approval pathways based on standard benchmarks may inadvertently authorize systems that are unsafe for long-tail patient populations \cite{weidinger2021ethical}.

\section{Open Challenges and Future Directions}

The transition of LLMs from experimental artifacts to core information infrastructure necessitates a rigorous accounting of their limitations. Our paper indicates that the LTK problem is not merely a technical error term to be minimized but a structural feature of learning from heavy-tailed distributions. This section outlines critical unresolved challenges and emerging future directions. 

\subsection{Measuring and Evaluating Long-Tail Knowledge in Practice}
A primary obstacle to progress is the lack of a stable operational definition for LTK. Current definitions largely rely on frequency counts in pre-training corpora such as Common Crawl, which fail to capture domain-specific importance or informational density \cite{kandpal2023struggle}. What constitutes a rare fact in a general web corpus may be foundational in specialized domains such as materials science or contract law. When benchmarks rely on aggregate metrics they compress heterogeneous failure modes into a single number, obscuring performance on the tail. Raji et al. \cite{raji2022fallacy} describe this as the fallacy of AI functionality, where aggregate success is mistaken for universal competence.

Evaluation validity is further compromised by static benchmarks. As models scale, overlap between test sets and training data increases, blurring the line between generalization and memorization \cite{sainz2023nlp, jacovi2023stop}. HELM-style evaluations demonstrate that performance is highly sensitive to prompt formulation, with variance most pronounced for rare knowledge lacking robust internal representations \cite{liang2023holistic, mishra2022reframing}. Consequently, the field lacks standardized methods to certify long-tail competence without costly manual audits \cite{widder2023limits}.

These findings suggest a need for tail-aware evaluation practices that explicitly stratify performance by frequency, domain criticality, or representational importance. Future work must reconsider whether benchmark-centric evaluation is sufficient for systems deployed in settings where tail failures dominate risk.

\subsection{Accountability Under Knowledge Uncertainty}
The probabilistic nature of long-tail failures complicates existing accountability frameworks. Unlike conventional software errors, LLM failures in the tail are stochastic and context-dependent. A model may retrieve a rare fact correctly in one context but hallucinate in another following minor prompt perturbations \cite{turpin2023unfaithful}. This instability diffuses responsibility across data curation, model architecture, deployment choices, and user prompting strategies. Failures that occur only in sparse regions of the input space may evade regulatory scrutiny despite being structurally predictable \cite{kroll2021accountable, casper2024blackbox}.

A key future direction is the development of accountability frameworks that explicitly recognize knowledge uncertainty. Rather than attributing errors solely post hoc, governance approaches may need to incorporate documented coverage limits, scoped use policies, or uncertainty-aware deployment standards.

\subsection{Privacy and Sustainability Constraints on Long-Tail Learning}
Efforts to improve long-tail performance face structural constraints related to privacy and sustainability. Long-tail examples are disproportionately vulnerable to memorization and extraction attacks. Carlini et al. \cite{carlini2021extracting} show that rare training examples are most susceptible to verbatim leakage. Techniques such as up-sampling or aggressive deduplication can improve rare fact retention but simultaneously increase the risk of exposing sensitive personally identifiable information \cite{brown2022does, kandpal2022deduplication}.

Scaling laws imply that capturing increasingly rare knowledge requires exponentially larger models and datasets. The environmental cost of training large dense models is well documented \cite{strubell2019energy, patterson2021carbon}. Although sparse architectures such as Mixture-of-Experts offer efficiency gains, the marginal cost of learning the next rarest fact continues to rise. This raises fundamental questions about the economic and ecological viability of encoding the full breadth of human knowledge into parametric models \cite{bender2021parrots, thompson2020computational}.

Future research must grapple with whether long-tail robustness should be pursued through parametric memorization at all, or whether responsibility for rare knowledge should increasingly shift to external, governed knowledge infrastructures.

\subsection{Long-Tail Knowledge as a Dynamic Sociotechnical Phenomenon}
The long tail is not static. It evolves through interaction between models, users, and information ecosystems. Deployment creates feedback loops that reshape knowledge distributions. Large-scale use of model-generated content reduces informational diversity, a process that can disproportionately erodes rare knowledge \cite{shumailov2024curse}. If future models are trained on AI-generated data, omissions and distortions of rare facts may become permanent.

Reliance on retrieval-augmented systems introduces additional dependencies. The availability of LTK becomes contingent on external search indices, which are themselves shaped by commercial incentives and algorithmic bias \cite{kulshrestha2019search}. Changes in indexing or access policies can instantaneously alter a model’s effective knowledge base. This shifts the locus of inquiry from model parameters to the broader sociotechnical ecosystem in which models operate \cite{selbst2019fairness}.

These dynamics suggest that preserving LTK is not solely a modeling problem but an ongoing governance challenge. Future work must treat long-tail robustness as a property of evolving sociotechnical systems rather than a static optimization target.

\section{Conclusion}
This paper studied the emerging body of research on LTK in LLMs, with the goal of clarifying how rare, sparse, and underrepresented information is defined, evaluated, and affected across the model lifecycle. Rather than treating long-tail failures as isolated anomalies, the studied literature consistently indicates that such failures arise from systematic interactions between data distributions, training objectives, representational capacity, and inference-time constraints. As LLMs are increasingly deployed as general-purpose knowledge systems, these limitations have become both technically consequential and sociotechnically salient.

We organized prior work along four complementary dimensions. First, we synthesized existing definitions of LTK into a taxonomy that captures linguistic, cultural, domain-specific, and temporal forms of sparsity, extending beyond purely frequency-based characterizations. Second, we reviewed mechanisms through which LTK is lost or weakly retained, including gradient dilution, representational interference, tokenization effects, and post-training compression. Third, we surveyed technical interventions proposed to mitigate long-tail failures, such as data-centric rebalancing, retrieval-augmented generation, architectural modularization, and model editing, highlighting their empirical scope and known tradeoffs. Finally, we examined sociotechnical implications, documenting how LTK gaps manifest as unequal system reliability, accountability ambiguity, trust miscalibration, and long-term feedback loops that further marginalize already underrepresented knowledge.

A key takeaway from this paper is that LTK failures cannot be understood or addressed in isolation at a single layer of the system. Improvements in model scale or architecture do not fully compensate for sparsity in training data, while inference-time augmentation alone does not resolve deeper issues of representation and evaluation. Conversely, many sociotechnical risks associated with epistemic exclusion and uneven system performance are directly linked to technical design choices that prioritize average-case accuracy over coverage of the tail. This interdependence suggests that assessments of LLM reliability, fairness, and accountability must explicitly account for long-tail behavior rather than treating it as residual error.

This paper contributes a structured synthesis that maps how disparate strands of research relate to one another and where key gaps remain. By clarifying definitions, mechanisms, evaluation practices, and implications within a unified analytical framework, we aim to support more systematic empirical studies and more transparent assessments of LLM capabilities and limitations.

As LLMs continue to mediate access to information across languages, domains, and communities, understanding what these systems systematically fail to know is as important as measuring what they perform well on. LTK will remain a central challenge for both technical robustness and sociotechnical responsibility, and future work will benefit from evaluation and deployment practices that make these limitations visible rather than implicit.

\section{Generative AI Usage Statement}

All study design, literature review, synthesis, and writing were conducted by the authors. Generative AI tools (Gemini) were used only for grammar checking and proofreading during the final polishing of the manuscript. No generative AI system was used to generate content, interpret prior work, or draw conclusions. The authors reviewed and approved all final text and remain fully responsible for the content of the paper.

\bibliographystyle{ACM-Reference-Format}
\bibliography{long_tail_bib2}

\appendix

\end{document}